\documentclass[11pt]{article}
\usepackage{acl2015}
\usepackage{times}
\usepackage{url}
\usepackage{latexsym}
\usepackage{graphicx}
\usepackage{multirow}
\usepackage{comment}
\usepackage{booktabs}
\usepackage{amsmath,amsfonts,amssymb}
\pdfpagewidth=\paperwidth
\pdfpageheight=\paperheight
\DeclareMathOperator*{\argmax}{arg\,max}

\title{Neural CRF Parsing}

\author{Greg Durrett \and Dan Klein\\
  Computer Science Division \\
  University of California, Berkeley \\
  {\tt \{gdurrett,klein\}@cs.berkeley.edu}\\
}

\date{}

\begin{document}
\maketitle
\begin{abstract}
This paper describes a parsing model that combines the exact dynamic programming of CRF parsing with the rich nonlinear featurization of neural net approaches. Our model is structurally a CRF that factors over anchored rule productions, but instead of linear potential functions based on sparse features, we use nonlinear potentials computed via a feedforward neural network. Because potentials are still local to anchored rules, structured inference (CKY) is unchanged from the sparse case. Computing gradients during learning involves backpropagating an error signal formed from standard CRF sufficient statistics (expected rule counts). Using only dense features, our neural CRF already exceeds a strong baseline CRF model \cite{HallEtAl2014}. In combination with sparse features, our system\footnote{System available at {\fontsize{2.85mm}{1em} \texttt{http://nlp.cs.berkeley.edu}}} achieves 91.1 F$_1$ on section 23 of the Penn Treebank, and more generally outperforms the best prior single parser results on a range of languages.
\end{abstract}

\section{Introduction}

Neural network-based approaches to structured NLP tasks have both strengths and weaknesses when compared to more conventional models, such conditional random fields (CRFs). A key strength of neural approaches is their ability to learn nonlinear interactions between underlying features.  In the case of unstructured output spaces, this capability has led to gains in problems ranging from syntax \cite{ChenManning2014,BelinkovEtAl2014} to lexical semantics \cite{KalchbrennerEtAl2014,Kim2014}. Neural methods are also powerful tools in the case of structured output spaces. Here, past work has often relied on recurrent architectures \cite{Henderson2003,SocherEtAl2013a,IrsoyCardie2014}, which can propagate information through structure via real-valued hidden state, but as a result do not admit efficient dynamic programming \cite{SocherEtAl2013a,LeZuidema2014}. However, there is a natural marriage of nonlinear induced features and efficient structured inference, as explored by \newcite{CollobertEtAl2011} for the case of sequence modeling: feedforward neural networks can be used to score local decisions which are then ``reconciled'' in a discrete structured modeling framework, allowing inference via dynamic programming.

In this work, we present a CRF constituency parser based on these principles, where individual anchored rule productions are scored based on nonlinear features computed with a feedforward neural network. A separate, identically-parameterized replicate of the network exists for each possible span and split point. As input, it takes vector representations of words at the split point and span boundaries; it then outputs scores for anchored rules applied to that span and split point. These scores can be thought of as nonlinear potentials analogous to linear potentials in conventional CRFs. Crucially, while the network replicates are connected in a unified model, their computations factor along the same substructures as in standard CRFs.

Prior work on parsing using neural network models has often sidestepped the problem of structured inference by making sequential decisions \cite{Henderson2003,ChenManning2014,Tsuboi2014} or by doing reranking \cite{SocherEtAl2013a,LeZuidema2014}; by contrast, our framework permits exact inference via CKY, since the model's structured interactions are purely discrete and do not involve continuous hidden state. Therefore, we can exploit a neural net's capacity to learn nonlinear features without modifying our core inference mechanism, allowing us to use tricks like coarse pruning that make inference efficient in the purely sparse model. Our model can be trained by gradient descent exactly as in a conventional CRF, with the gradient of the network parameters naturally computed by backpropagating a difference of expected anchored rule counts through the network for each span and split point.


Using dense learned features alone, the neural CRF model obtains high performance, outperforming the CRF parser of \newcite{HallEtAl2014}. When sparse indicators are used in addition, the resulting model gets 91.1 F$_1$ on section 23 of the Penn Treebank, outperforming the parser of \newcite{SocherEtAl2013a} as well as the Berkeley Parser \cite{PetrovKlein2007} and matching the discriminative parser of \newcite{CarrerasEtAl2008}. The model also obtains the best single parser results on nine other languages, again outperforming the system of \newcite{HallEtAl2014}.

\section{Model}

\begin{figure}[t!]
\begin{centering}
\includegraphics[trim=2mm 68mm 55mm 0mm,scale=0.345]{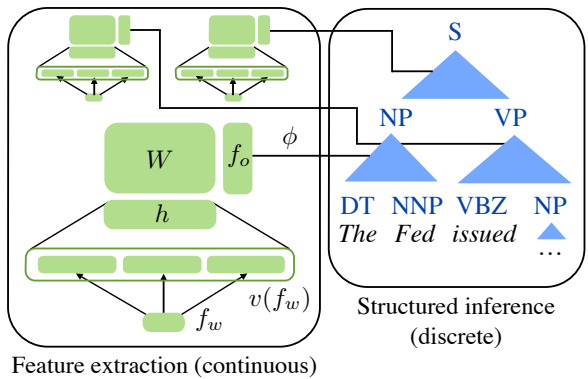}
\caption{\label{fig:model} Neural CRF model. On the right, each anchored rule $(r,s)$ in the tree is independently scored by a function $\phi$, so we can perform inference with CKY to compute marginals or the Viterbi tree. On the left, we show the process for scoring an anchored rule with neural features: words in $f_w$ (see Figure~\ref{fig:example}) are embedded, then fed through a neural network with one hidden layer to compute dense intermediate features, whose conjunctions with sparse rule indicator features $f_o$ are scored according to parameters $W$.
}
\end{centering}
\end{figure}

Figure~\ref{fig:model} shows our neural CRF model. The model decomposes over anchored rules, and it scores each of these with a potential function; in a standard CRF, these potentials are typically linear functions of sparse indicator features, whereas in our approach they are nonlinear functions of word embeddings.\footnote{Throughout this work, we will primarily consider two potential functions: linear functions of sparse indicators and nonlinear neural networks over dense, continuous features. Although other modeling choices are possible, these two points in the design space reflect common choices in NLP, and past work has suggested that nonlinear functions of indicators or linear functions of dense features may perform less well \cite{WangManning2013}.} Section~\ref{sec:anchored} describes our notation for anchored rules, and Section~\ref{sec:scoring} talks about how they are scored. We then discuss specific choices of our featurization (Section~\ref{sec:features}) and the backbone grammar used for structured inference (Section~\ref{sec:grammar}).

\begin{figure}[t!]
\begin{centering}
\includegraphics[trim=13mm 50mm 55mm 00mm,scale=0.33]{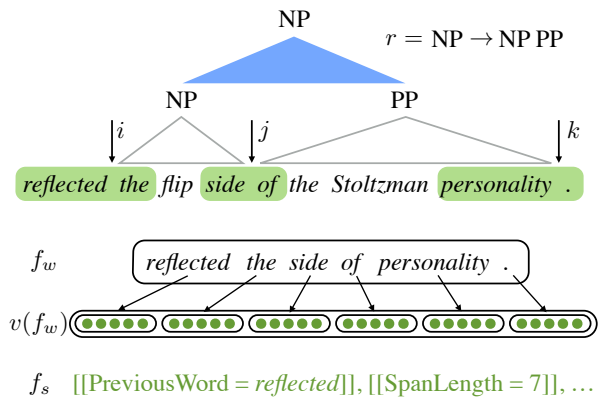}
\caption{\label{fig:example} Example of an anchored rule production for the rule NP $\rightarrow$ NP PP. From the anchoring $s=(i,j,k)$, we extract either sparse surface features $f_s$ or a sequence of word indicators $f_w$ which are embedded to form a vector representation $v(f_w)$ of the anchoring's lexical properties.
}
\end{centering}
\end{figure}

\subsection{Anchored Rules}
\label{sec:anchored}

The fundamental units that our parsing models consider are \emph{anchored rules}. As shown in Figure~\ref{fig:example}, we define an anchored rule as a tuple $(r,s)$, where $r$ is an indicator of the rule's identity and $s = (i,j,k)$ indicates the span $(i,k)$ and split point $j$ of the rule.\footnote{For simplicity of exposition, we ignore unary rules; however, they are easily supported in this framework by simply specifying a null value for the split point.} A tree $T$ is simply a collection of anchored rules subject to the constraint that those rules form a tree. All of our parsing models are CRFs that decompose over anchored rule productions and place a probability distribution over trees conditioned on a sentence $\mathbf{w}$ as follows:
\begin{equation*}
P(T|\mathbf{w}) \propto \exp \left(\sum_{(r,s) \in T} \phi(\mathbf{w},r,s)\right)
\end{equation*}
where $\phi$ is a scoring function that considers the input sentence and the anchored rule in question. Figure~\ref{fig:model} shows this scoring process schematically. As we will see, the module on the left can be be a neural net, a linear function of surface features, or a combination of the two, as long as it provides anchored rule scores, and the structured inference component is the same regardless (CKY).

A PCFG estimated with maximum likelihood has $\phi(\mathbf{w},r,s) = \log P(r|\textrm{parent}(r))$, which is independent of the anchoring $s$ and the words $\mathbf{w}$ except for preterminal productions; a basic discriminative parser might let this be a learned parameter but still disregard the surface information. However, surface features can capture useful syntactic cues \cite{FinkelEtAl2008,HallEtAl2014}. Consider the example in Figure~\ref{fig:example}: the proposed parent NP is preceded by the word \emph{reflected} and followed by a period, which is a surface context characteristic of NPs or PPs in object position. Beginning with \emph{the} and ending with \emph{personality} are typical properties of NPs as well, and the choice of the particular rule NP $\rightarrow$ NP PP is supported by the fact that the proposed child PP begins with \emph{of}. This information can be captured with sparse features ($f_s$ in Figure~\ref{fig:example}) or, as we describe below, with a neural network taking lexical context as input.

\subsection{Scoring Anchored Rules}
\label{sec:scoring}

\begin{figure}[t!]
\begin{centering}
\includegraphics[trim=12mm 70mm 55mm 10mm,scale=0.35]{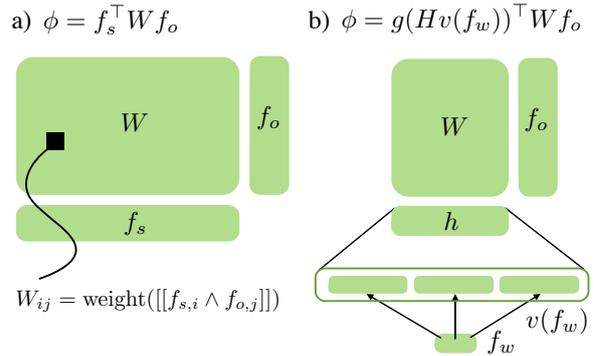}
\caption{\label{fig:scoring} Our sparse (left) and neural (right) scoring functions for CRF parsing. $f_s$ and $f_w$ are raw surface feature vectors for the sparse and neural models (respectively) extracted over anchored spans with split points. (a) In the sparse case, we multiply $f_s$ by a weight matrix $W$ and then a sparse output vector $f_o$ to score the rule production. (b) In the neural case, we first embed $f_w$ and then transform it with a one-layer neural network in order to produce an intermediate feature representation $h$ before combining with $W$ and $f_o$.
}
\end{centering}
\end{figure}

Following \newcite{HallEtAl2014}, our baseline sparse scoring function takes the following bilinear form:
\begin{equation*}
\phi_{\textrm{sparse}}(\mathbf{w},r,s;W) = f_s(\mathbf{w},s)^\top W f_o(r)
\end{equation*}
where $f_o(r) \in \{0,1\}^{n_o}$ is a sparse vector of features expressing properties of $r$ (such as the rule's identity or its parent label) and $f_s(\mathbf{w},s) \in \{0,1\}^{n_s}$ is a sparse vector of surface features associated with the words in the sentence and the anchoring, as shown in Figure~\ref{fig:example}. $W$ is a $n_s \times n_o$ matrix of weights.\footnote{A more conventional expression of the scoring function for a CRF is $\phi(\mathbf{w},r,s)\mkern-3mu = \theta^{\top}\mkern-4mu f(\mathbf{w},r,s)$, with a vector $\theta$ for the parameters and a single feature extractor $f$ that jointly inspects the surface and the rule. However, when the feature representation conjoins each rule $r$ with surface properties of the sentence in a systematic way (an assumption that holds in our case as well as for standard CRF models for POS tagging and NER), this is equivalent to our formalism.} The scoring of a particular anchored rule is depicted in Figure~\ref{fig:scoring}a; note that surface features and rule indicators are conjoined in a systematic way.

The role of $f_s$ can be equally well played by a vector of dense features learned via a neural network. We will now describe how to compute these features, which represent a transformation of surface lexical indicators $f_w$. Define $f_w(\mathbf{w},s) \in \mathbb{N}^{n_w}$ to be a function that produces a fixed-length sequence of word indicators based on the input sentence and the anchoring. This vector of word identities is then passed to an embedding function $v: \mathbb{N} \rightarrow \mathbb{R}^{n_e}$ and the dense representations of the words are subsequently concatenated to form a vector we denote by $v(f_w)$.\footnote{Embedding words allows us to use standard pre-trained vectors more easily and tying embeddings across word positions substantially reduces the number of model parameters. However, embedding features rather than words has also been shown to be effective \cite{ChenEtAl2014}.} Finally, we multiply this by a matrix $H \in \mathbb{R}^{n_h \times (n_w n_e)}$ of real-valued parameters and pass it through an elementwise nonlinearity $g(\cdot)$. We use rectified linear units $g(x) = \max(x,0)$ and discuss this choice more in Section~\ref{sec:design}.

Replacing $f_s$ with the end result of this computation $h(\mathbf{w},s;H) = g(H v(f_w (\mathbf{w},s)))$, our scoring function becomes
\begin{equation*}
\phi_{\textrm{neural}}(\mathbf{w},r,s;H,W) = h(\mathbf{w},s;H)^\top W f_o(r)
\end{equation*}
as shown in Figure~\ref{fig:scoring}b. For a fixed $H$, this model can be viewed as a basic CRF with dense input features. By learning $H$, we learn intermediate feature representations that provide the model with more discriminating power. Also note that it is possible to use deeper networks or more sophisticated architectures here; we will return to this in Section~\ref{sec:design}.

Our two models can be easily combined:
\begin{align*}
\phi(\mathbf{w},r,s;W_1,H,W_2) &= \phi_{\textrm{sparse}}(\mathbf{w},r,s;W_1)\\
&+ \phi_{\textrm{neural}}(\mathbf{w},r,s;H,W_2)
\end{align*}
Weights for each component of the scoring function can be learned fully jointly and inference proceeds as before.

\subsection{Features}
\label{sec:features}

We take $f_s$ to be the set of features described in \newcite{HallEtAl2014}. At the preterminal layer, the model considers prefixes and suffixes up to length 5 of the current word and neighboring words, as well as the words' identities. For nonterminal productions, we fire indicators on the words\footnote{The model actually uses the longest suffix of each word occurring at least 100 times in the training set, up to the entire word. Removing this abstraction of rare words harms performance.} before and after the start, end, and split point of the anchored rule (as shown in Figure~\ref{fig:example}) as well as on two other span properties, span length and span shape (an indicator of where capitalized words, numbers, and punctuation occur in the span).

For our neural model, we take $f_w$ for all productions (preterminal and nonterminal) to be the words surrounding the beginning and end of a span and the split point, as shown in Figure~\ref{fig:example}; in particular, we look two words in either direction around each point of interest, meaning the neural net takes 12 words as input.\footnote{The sparse model did not benefit from using this larger neighborhood, so improvements from the neural net are not simply due to considering more lexical context.} For our word embeddings $v$, we use pre-trained word vectors from \newcite{BansalEtAl2014}. We compare with other sources of word vectors in Section~\ref{sec:dev}. Contrary to standard practice, we do not update these vectors during training; we found that doing so did not provide an accuracy benefit and slowed down training considerably.

\subsection{Grammar Refinements}
\label{sec:grammar}

A recurring issue in discriminative constituency parsing is the granularity of annotation in the base grammar \cite{FinkelEtAl2008,PetrovKlein2008,HallEtAl2014}. Using finer-grained symbols in our rules $r$ gives the model greater capacity, but also introduces more parameters into $W$ and increases the ability to overfit. Following \newcite{HallEtAl2014}, we use grammars with very little annotation: we use no horizontal Markovization for any of experiments, and all of our English experiments with the neural CRF use no vertical Markovization ($V=0$). This also has the benefit of making the system much faster, due to the smaller state space for dynamic programming. We do find that using parent annotation ($V=1$) is useful on other languages (see Section~\ref{sec:spmrl}), but this is the only grammar refinement we consider.

\section{Learning}
\label{sec:learning}

To learn weights for our neural model, we maximize the conditional log likelihood of our $D$ training trees $T^*$:
\begin{equation*}
\mathcal{L}(H,W) = \sum_{i=1}^D \log P(T^*_i|\mathbf{w}_i;H,W)
\end{equation*}
Because we are using rectified linear units as our nonlinearity, our objective is not everywhere differentiable. The interaction of the parameters and the nonlinearity also makes the objective non-convex. However, in spite of this, we can still follow subgradients to optimize this objective, as is standard practice.

Recall that $h(\mathbf{w},s;H)$ are the hidden layer activations. The gradient of $W$ takes the standard form of log-linear models:
\begin{align*}
&\frac{\partial \mathcal{L}}{\partial W} = \left(\sum_{(r,s) \in T^*} h(\mathbf{w},s;H) f_o(r)^\top\right) - \\
&\left(\sum_T P(T|\mathbf{w};H,W) \sum_{(r,s) \in T} h(\mathbf{w},s;H) f_o(r)^\top\right)
\end{align*}
Note that the outer products give matrices of feature counts isomorphic to $W$. The second expression can be simplified to be in terms of expected feature counts. To update $H$, we use standard backpropagation by first computing:
\begin{align*}
\frac{\partial \mathcal{L}}{\partial h} &= \left(\sum_{(r,s) \in T^*} W f_o(r) \right) - \\
& \left(\sum_T P(T|\mathbf{w};H,W) \sum_{(r,s) \in T} W f_o(r)\right)
\end{align*}
Since $h$ is the output of the neural network, we can then apply the chain rule to compute gradients for $H$ and any other parameters in the neural network. 

Learning uses Adadelta \cite{Zeiler2014}, which has been employed in past work \cite{Kim2014}. We found that Adagrad \cite{DuchiEtAl2011} performed equally well with tuned regularization and step size parameters, but Adadelta worked better out of the box. We set the momentum term $\rho = 0.95$ (as suggested by \newcite{Zeiler2014}) and did not regularize the weights at all. We used a minibatch size of 200 trees, although the system was not particularly sensitive to this. For each treebank, we trained for either 10 passes through the treebank or 1000 minibatches, whichever is shorter.

We initialized the output weight matrix $W$ to zero. To break symmetry, the lower level neural network parameters $H$ were initialized with each entry being independently sampled from a Gaussian with mean 0 and variance 0.01; Gaussian performed better than uniform initialization, but the variance was not important.

\section{Inference}

Our baseline and neural model both score anchored rule productions. We can use CKY in the standard fashion to compute either expected anchored rule counts $\mathbb{E}_{P(T|\mathbf{w})}[(r,s)]$ or the Viterbi tree $\argmax_T P(T|\mathbf{w})$.

We speed up inference by using a coarse pruning pass. We follow \newcite{HallEtAl2014} and prune according to an X-bar grammar with head-outward binarization, ruling out any constituent whose max marginal probability is less than $e^{-9}$. With this pruning, the number of spans and split points to be considered is greatly reduced; however, we still need to compute the neural network activations for each remaining span and split point, of which there may be thousands for a given sentence.\footnote{One reason we did not choose to include the rule identity $f_o$ as an input to the network is that it requires computing an even larger number of network activations, since we cannot reuse them across rules over the same span and split point.} We can improve efficiency further by noting that the same word will appear in the same position in a large number of span/split point combinations, and cache the contribution to the hidden layer caused by that word \cite{ChenManning2014}. Computing the hidden layer then simply requires adding $n_w$ vectors together and applying the nonlinearity, instead of a more costly matrix multiply.

Because the number of rule indicators $n_o$ is fairly large (approximately 4000 in the Penn Treebank), the multiplication by $W$ in the model is also expensive. However, because only a small number of rules can apply to a given span and split point, $f_o$ is sparse and we can selectively compute the terms necessary for the final bilinear product.


Our combined sparse and neural model trains on the Penn Treebank in 24 hours on a single machine with a parallelized CPU implementation. For reference, the purely sparse model with a parent-annotated grammar (necessary for the best results) takes around 15 hours on the same machine.

\section{System Ablations}
\label{sec:dev}

Table~\ref{table:dev_results} shows results on section 22 (the development set) of the English Penn Treebank \cite{MarcusEtAl1993}, computed using \texttt{evalb}. Full test results and comparisons to other systems are shown in Table~\ref{table:test_results}. We compare variants of our system along two axes: whether they use standard linear sparse features, nonlinear dense features from the neural net, or both, and whether any word representations (vectors or clusters) are used.

\paragraph{Sparse vs. neural} The neural CRF (line (d) in Table~\ref{table:dev_results}) on its own outperforms the sparse CRF (a, b) even when the sparse CRF has a more heavily annotated grammar. This is a surprising result: the features in the sparse CRF have been carefully engineered to capture a range of linguistic phenomena \cite{HallEtAl2014}, and there is no guarantee that word vectors will capture the same. For example, at the POS tagging layer, the sparse model looks at prefixes and suffixes of words, which give the model access to morphology for predicting tags of unknown words, which typically have regular inflection patterns. By contrast, the neural model must rely on the geometry of the vector space exposing useful regularities. At the same time, the strong performance of the combination of the two systems (g) indicates that not only are both featurization approaches high-performing on their own, but that they have complementary strengths.

\paragraph{Unlabeled data} Much attention has been paid to the choice of word vectors for various NLP tasks, notably whether they capture more syntactic or semantic phenomena \cite{BansalEtAl2014,LevyGoldberg2014}. We primarily use vectors from \newcite{BansalEtAl2014}, who train the skip-gram model of \newcite{MikolovEtAl2013} using contexts from dependency links; a similar approach was also suggested by \newcite{LevyGoldberg2014}. However, as these embeddings are trained on a relatively small corpus (BLLIP minus the Penn Treebank), it is natural to wonder whether less-syntactic embeddings trained on a larger corpus might be more useful. This is not the case: line (e) in Table~\ref{table:dev_results} shows the performance of the neural CRF using the Wikipedia-trained word embeddings of \newcite{CollobertEtAl2011}, which  do not perform better than the vectors of \newcite{BansalEtAl2014}.

\begin{table}[t]
\begin{center}
\small
\renewcommand{\tabcolsep}{1.3mm}
\begin{tabular}{ccccccc} \toprule
 & Sparse & Neural & $V$ & Word Reps & F$_1$ len $\leq 40$ & F$_1$ all \\ \midrule
 \multicolumn{5}{c}{\newcite{HallEtAl2014}, $V=1$} & 90.5\phantom{0} &  \\ \midrule
a & \checkmark &  & 0 & & 89.89 & 89.22 \\
b & \checkmark &  & 1 & & 90.82 & 90.13 \\
c & \checkmark &  & 1 & Brown & 90.80 & 90.17 \\ \midrule
d & & \checkmark  & 0 & Bansal & 90.97 & 90.44 \\
e & & \checkmark & 0 & Collobert & 90.25 & 89.63 \\
f & & \checkmark & 0 & PTB & 89.34 & 88.99 \\ \midrule
g & \checkmark & \checkmark & 0 & Bansal & \textbf{92.04} & \textbf{91.34} \\
h & \checkmark & \checkmark & 0 & PTB & 91.39 & 90.91 \\ \bottomrule
\end{tabular}
\end{center}
\caption{\label{table:dev_results} Results of our sparse CRF, neural CRF, and combined parsing models on section 22 of the Penn Treebank. Systems are broken down by whether local potentials come from sparse features and/or the neural network (the primary contribution of this work), their level of vertical Markovization, and what kind of word representations they use. The neural CRF (d) outperforms the sparse CRF (a, b) even when a more heavily annotated grammar is used, and the combined approach (g) is substantially better than either individual model. The contribution of the neural architecture cannot be replaced by Brown clusters (c), and even word representations learned just on the Penn Treebank are surprisingly effective (f, h).
}
\end{table}

To isolate the contribution of continuous word representations themselves, we also experimented with vectors trained on just the text from the training set of the Penn Treebank using the skip-gram model with a window size of 1. While these vectors are somewhat lower performing on their own (f), they still provide a surprising and noticeable gain when stacked on top of sparse features (h), again suggesting that dense and sparse representations have complementary strengths. This result also reinforces the notion that the utility of word vectors does \emph{not} come primarily from importing information about out-of-vocabulary words \cite{AndreasKlein2014}.

Since the neural features incorporate information from unlabeled data, we should provide the sparse model with similar information for a true apples-to-apples comparison. Brown clusters have been shown to be effective vehicles in the past \cite{KooEtAl2008,TurianEtAl2010,BansalEtAl2014}. We can incorporate Brown clusters into the baseline CRF model in an analogous way to how embedding features are used in the dense model: surface features are fired on Brown cluster identities (we use prefixes of length 4 and 10) of key words. We use the Brown clusters from \newcite{KooEtAl2008}, which are trained on the same data as the vectors of \newcite{BansalEtAl2014}. However, Table~\ref{table:dev_results} shows that these features provide no benefit to the baseline model, which suggests either that it is difficult to learn reliable weights for these as sparse features or that different regularities are being captured by the word embeddings.

\section{Design Choices}
\label{sec:design}

The neural net design space is large, so we wish to analyze the particular design choices we made for this system by examining the performance of several variants of the neural net architecture used in our system. Table~\ref{table:neural_ablations} shows development results from potential alternate architectural choices, which we now discuss.

\begin{table}[t]
\begin{center}
\begin{tabular}{rrcc} \toprule
 & & F$_1$ len $\leq 40$ & $\Delta$\\ \midrule
\multicolumn{2}{c}{Neural CRF}  & 90.97        & ---\\ \midrule
\multirow{3}{*}{Nonlinearity} &
ReLU   & 90.97          & --- \\ 
& Tanh   & 90.74        & $-0.23$\\ 
& Cube   & 89.94        & $-1.03$\\ \midrule
\multirow{3}{*}{Depth}
& 0 HL   & 90.54        & $-0.43$\\ 
& 1 HL   & 90.97        & ---\\ 
& 2 HL   & 90.58        & $-0.39$\\ \midrule
\multicolumn{2}{c}{Embed output} & 88.81  & $-2.16$\\ \midrule
\end{tabular}
\end{center}
\caption{\label{table:neural_ablations} Exploration of other implementation choices in the feedforward neural network on sentences of length $\leq 40$ from section 22 of the Penn Treebank. Rectified linear units perform better than tanh or cubic units, a network with one hidden layer performs best, and embedding the output feature vector gives worse performance. 
}
\end{table}

\paragraph{Choice of nonlinearity} The choice of nonlinearity $g$ has been frequently discussed in the neural network literature. Our choice $g(x) = \max(x,0)$, a rectified linear unit, is increasingly popular in computer vision \cite{KrizhevskyEtAl2012}. $g(x) = \tanh(x)$ is a traditional nonlinearity widely used throughout the history of neural nets \cite{BengioEtAl2003}. $g(x) = x^3$ (cube) was found to be most successful by \newcite{ChenManning2014}.

Table~\ref{table:neural_ablations} compares the performance of these three nonlinearities. We see that rectified linear units perform the best, followed by $\tanh$ units, followed by cubic units.\footnote{The performance of cube decreased substantially late in learning; it peaked at around 90.52. Dropout may be useful for alleviating this type of overfitting, but in our experiments we did not find dropout to be beneficial overall.} One drawback of $\tanh$ as an activation function is that it is easily ``saturated'' if the input to the unit is too far away from zero, causing the backpropagation of derivatives through that unit to essentially cease; this is known to cause problems for training, requiring special purpose machinery for use in deep networks \cite{IoffeSzegedy2015}.

\begin{figure}[t!]
\begin{centering}
\includegraphics[trim=35mm 66mm 55mm 10mm,scale=0.31]{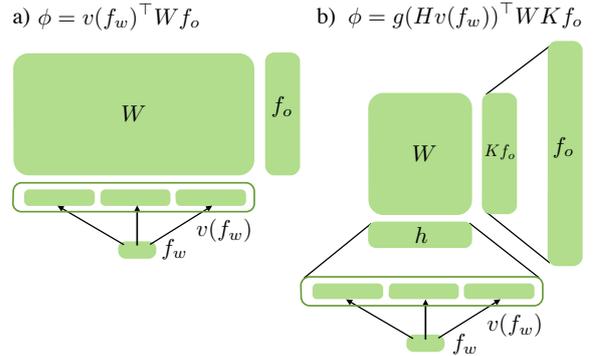}
\caption{\label{fig:model-linoe} Two additional forms of the scoring function. a) Linear version of the dense model, equivalent to a CRF with continuous-valued input features. b) Version of the dense model where outputs are also embedded according to a learned matrix $K$.
}
\end{centering}
\end{figure}

\paragraph{Depth} Given that we are using rectified linear units, it bears asking whether or not our implementation is improving substantially over linear features of the continuous input. We can use the embedding vector of an anchored span $v(f_w)$ directly as input to a basic linear CRF, as shown in Figure~\ref{fig:model-linoe}a. Table~\ref{table:dev_results} shows that the purely linear architecture (0 HL) performs surprisingly well, but is still less effective than the network with one hidden layer.  This agrees with the results of \newcite{WangManning2013}, who noted that dense features typically benefit from nonlinear modeling. We also compare against a two-layer neural network, but find that this also performs worse than the one-layer architecture.

\paragraph{Densifying output features} Overall, it appears beneficial to use dense representations of surface features; a natural question that one might ask is whether the same technique can be applied to the sparse output feature vector $f_o$. We can apply the approach of \newcite{SrikumarManning2014} and multiply the sparse output vector by a dense matrix $K$, giving the following scoring function (shown in Figure~\ref{fig:model-linoe}b):
\begin{equation*}
\resizebox{1.0\hsize}{!}{$\phi(\mathbf{w},r,s;H,W,K) = g(H v(f_w(\mathbf{w},s)))^\top WK f_o(r)$}
\end{equation*}
where $W$ is now $n_h \times n_{oe}$ and $K$ is $n_{oe} \times n_o$. $WK$ can be seen a low-rank approximation of the original $W$ at the output layer, similar to low-rank factorizations of parameter matrices used in past work \cite{LeiEtAl2014}. This approach saves us from having to learn a separate row of $W$ for every rule in the grammar; if rules are given similar embeddings, then they will behave similarly according to the model.


We experimented with $n_{oe} = 20$ and show the results in Table~\ref{table:neural_ablations}. Unfortunately, this approach does not seem to work well for parsing. Learning the output representation was empirically very unstable, and it also required careful initialization. We tried Gaussian initialization (as in the rest of our model) and initializing the model by clustering rules either randomly or according to their parent symbol. The latter is what is shown in the table, and gave substantially better performance. We hypothesize that blurring distinctions between output classes may harm the model's ability to differentiate between closely-related symbols, which is required for good parsing performance. Using pre-trained rule embeddings at this layer might also improve performance of this method.

\begin{table*}[t!]
\small
\renewcommand{\tabcolsep}{1.4mm}
\centering
\begin{tabular}{rcccccccccc}
\toprule
            & Arabic & Basque & French & German & Hebrew & Hungarian & Korean & Polish & Swedish & Avg \\ \midrule
 \multicolumn{11}{c}{Dev, all lengths} \\\midrule
\newcite{HallEtAl2014} & 78.89 & 83.74 & 79.40 & 83.28 & 88.06 & 87.44 & 81.85 & 91.10 & 75.95 & 83.30 \\ 
This work*              & \textbf{80.68} & \textbf{84.37} & \textbf{80.65} & \textbf{85.25} & \textbf{89.37} & \textbf{89.46} & \textbf{82.35} & \textbf{92.10} & \textbf{77.93} & \textbf{84.68} \\ \toprule\toprule
 \multicolumn{11}{c}{Test, all lengths} \\\toprule
Berkeley               & 79.19 & 70.50 & 80.38 & 78.30 & 86.96 & 81.62 & 71.42 & 79.23 & 79.18 & 78.53\\
Berkeley-Tags          & 78.66 & 74.74 & 79.76 & 78.28 & 85.42 & 85.22 & 78.56 & 86.75 & 80.64 & 80.89\\ 
\newcite{CrabbeSeddah2014}             & 77.66 & 85.35 & 79.68 & 77.15 & 86.19 & 87.51 & 79.35 & 91.60 & 82.72 & 83.02 \\
\newcite{HallEtAl2014} & 78.75 & 83.39 & 79.70 & 78.43 & 87.18 & 88.25 & 80.18 & 90.66 & 82.00 & 83.17\\ 
This work*              & \textbf{80.24} & \textbf{85.41} & \textbf{81.25} & \textbf{80.95} & \textbf{88.61} & \textbf{90.66} & \textbf{82.23} & \textbf{92.97} & \textbf{83.45} & \textbf{85.08} \\ \midrule
\multicolumn{11}{c}{Reranked ensemble} \\ \toprule
2014 Best      & \textbf{81.32} & \textbf{88.24} & \textbf{82.53} & \textbf{81.66} & \textbf{89.80} & \textbf{91.72} & \textbf{83.81} & \textbf{90.50} & \textbf{85.50} & \textbf{86.12}\\ \bottomrule
\end{tabular}
\caption{\label{table:spmrl} Results for the nine treebanks in the SPMRL 2013/2014 Shared Tasks; all values are F-scores for sentences of all lengths using the version of \texttt{evalb} distributed with the shared task. Our parser substantially outperforms the strongest single parser results on this dataset \cite{HallEtAl2014,CrabbeSeddah2014}. Berkeley-Tags is an improved version of the Berkeley parser designed for the shared task \cite{SeddahEtAl2013}. 2014 Best is a reranked ensemble of modified Berkeley parsers and constitutes the best published numbers on this dataset \cite{BjorkelundEtAl2013,BjorkelundEtAl2014}. 
}
\end{table*}

\begin{table}[h!]
\begin{center}
\begin{tabular}{rc} \toprule
 & F$_1$ all \\ \midrule
\multicolumn{2}{c}{Single model, PTB only} \\ \midrule 
\newcite{HallEtAl2014} & 89.2 \\
Berkeley  & 90.1 \\
\newcite{CarrerasEtAl2008} & 91.1 \\
\newcite{ShindoEtAl2012} single & 91.1 \\ \midrule
\multicolumn{2}{c}{Single model, PTB + vectors/clusters} \\ \midrule
\newcite{ZhuEtAl2013}      & 91.3 \\
This work* & 91.1 \\ \midrule
\multicolumn{2}{c}{Extended conditions} \\ \midrule
\newcite{CharniakJohnson2005} & 91.5 \\
\newcite{SocherEtAl2013a}    & 90.4 \\
\newcite{VinyalsEtAl2014} single & 90.5 \\
\newcite{VinyalsEtAl2014} ensemble & 91.6 \\
\newcite{ShindoEtAl2012} ensemble & 92.4 \\
\bottomrule
\end{tabular}
\end{center}
\caption{\label{table:test_results} Test results on section 23 of the Penn Treebank. We compare to several categories of parsers from the literatures. We outperform strong baselines such as the Berkeley Parser \cite{PetrovKlein2007} and the CVG Stanford parser \cite{SocherEtAl2013a} and we match the performance of sophisticated generative \cite{ShindoEtAl2012} and discriminative \cite{CarrerasEtAl2008} parsers.
}
\vspace{-0.1in}
\end{table}

\section{Test Results}
\label{sec:results}

We evaluate our system under two conditions: first, on the English Penn Treebank, and second, on the nine languages used in the SPMRL 2013 and 2014 shared tasks.

\subsection{Penn Treebank}

Table~\ref{table:test_results} reports results on section 23 of the Penn Treebank (PTB). We focus our comparison on single parser systems as opposed to rerankers, ensembles, or self-trained methods (though these are also mentioned for context). First, we compare against four parsers trained only on the PTB with no auxiliary data: the CRF parser of \newcite{HallEtAl2014}, the Berkeley parser \cite{PetrovKlein2007}, the discriminative parser of \newcite{CarrerasEtAl2008}, and the single TSG parser of \newcite{ShindoEtAl2012}. To our knowledge, the latter two systems are the highest performing in this PTB-only, single parser data condition; we match their performance at 91.1 F$_1$, though we also use word vectors computed from unlabeled data. We further compare to the shift-reduce parser of \newcite{ZhuEtAl2013}, which uses unlabeled data in the form of Brown clusters. Our method achieves performance close to that of their parser.

We also compare to the compositional vector grammar (CVG) parser of \newcite{SocherEtAl2013a} as well as the LSTM-based parser of \newcite{VinyalsEtAl2014}. The conditions these parsers are operating under are slightly different: the former is a reranker on top of the Stanford Parser \cite{KleinManning2003} and the latter trains on much larger amounts of data parsed by a product of Berkeley parsers \cite{PetrovKlein2010}. Regardless, we outperform the CVG parser as well as the single parser results from \newcite{VinyalsEtAl2014}.

\subsection{SPMRL}
\label{sec:spmrl}

We also examine the performance of our parser on other languages, specifically the nine morphologically-rich languages used in the SPMRL 2013/2014 shared tasks \cite{SeddahEtAl2013,SeddahEtAl2014}. We train word vectors on the monolingual data distributed with the SPMRL 2014 shared task (typically 100M-200M tokens per language) using the skip-gram approach of \texttt{word2vec} with a window size of 1 \cite{MikolovEtAl2013}.\footnote{Training vectors with the \textsc{Skip$_\textrm{DEP}$} method of \newcite{BansalEtAl2014} did not substantially improve performance here.} Here we use $V=1$ in the backbone grammar, which we found to be beneficial overall. Table~\ref{table:spmrl} shows that our system improves upon the performance of the parser from \newcite{HallEtAl2014} as well as the top single parser from the shared task \cite{CrabbeSeddah2014}, with robust improvements on all languages.


\section{Conclusion}

In this work, we presented a CRF parser that scores anchored rule productions using dense input features computed from a feedforward neural net. Because the neural component is modularized, we can easily integrate it into a pre-existing learning and inference framework based around dynamic programming of a discrete parse chart. Our combined neural and sparse model gives strong performance both on English and on other languages.

Our system is publicly available at \texttt{http://nlp.cs.berkeley.edu}.

\section*{Acknowledgments}
\label{sec:acknowledgments}

This work was partially supported by BBN under DARPA contract HR0011-12-C-0014, by a Facebook fellowship for the first author, and by a Google Faculty Research Award to the second author. Thanks to David Hall for assistance with the Epic parsing framework and for a preliminary implementation of the neural architecture, to Kush Rastogi for training word vectors on the SPMRL data, to Dan Jurafsky for helpful discussions, and to the anonymous reviewers for their insightful comments.

\bibliographystyle{acl}
\bibliography{nepic-final}

\end{document}